\documentclass[runningheads]{llncs}
\usepackage{graphicx}
\usepackage{comment}
\usepackage{amsmath,amssymb} 
\usepackage{color}

\usepackage[width=122mm,left=12mm,paperwidth=146mm,height=193mm,top=12mm,paperheight=217mm]{geometry}

\usepackage{color}
\usepackage{url}  
\usepackage{hyperref}      
\usepackage{booktabs}      
\usepackage{amsfonts}   
\usepackage{nicefrac}     
\usepackage{microtype}      
\usepackage{float}
\usepackage{makecell}
\usepackage{csquotes}

\begin{document}
\pagestyle{headings}
\mainmatter
\def\ECCVSubNumber{5394}  

\title{Multi-view adaptive graph convolutions for graph classification}

\titlerunning{Multi-view adaptive graph convolutions}

\author{Nikolas Adaloglou \and
Nicholas Vretos \and Petros Daras }
\authorrunning{Adaloglou et al.}
%
\institute{The Visual Computing Lab - Information Technologies Institute, Centre for Research and Technology Hellas, 57001 Thessaloniki, Greece\\ 
\email{\{adaloglou,vretos,daras\}@iti.gr}}
\maketitle

\begin{abstract}
In this paper, a novel multi-view methodology for graph-based neural networks is proposed. A systematic and methodological adaptation of the key concepts of classical deep learning methods such as convolution, pooling and multi-view architectures is developed for the context of non-Euclidean manifolds. The aim of the proposed work is to present a novel multi-view graph convolution layer, as well as a new view pooling layer making use of: a) a new hybrid Laplacian that is adjusted based on feature distance metric learning, b) multiple trainable representations of a feature matrix of a graph, using trainable distance matrices, adapting the notion of views to graphs and c) a multi-view graph aggregation scheme called graph view pooling, in order to synthesise information from the multiple generated \enquote{views}. The aforementioned layers are used in an end-to-end graph neural network architecture for graph classification and show competitive results to other state-of-the-art methods.
\end{abstract}

\section{Introduction}
Graph theory enabled us, among other things, to powerfully represent relationships between data. Due to their avast application field, ranging from biological structures to modern social networks, graph representations, along with the recently exploding field of Graph Neural Networks (GNNs), assisted us to re-address challenging graph related tasks under the prism of neural networks. To that end, the task of adapting operations from classical convolutional neural networks (CNNs), such as convolution and pooling, to non-Euclidean domains (e.g., graphs and manifolds), gave rise to the emerging field of geometric deep learning \cite{bronstein2017geometric}.

A typical CNN consists of different layers of convolutions and pooling that are usually combined in a sequential manner \cite{lecun1999object}, \cite{fukushima1980neocognitron}, \cite{krizhevsky2012imagenet}. The idea of adapting these key operations of CNNs to irregular grids is not new. Until recently, a considerable amount of research effort has been invested to adapt the classical convolution layers in many different ways towards graph-based convolutions \cite{bruna2013spectral} \nocite{henaff2015deep}\nocite{defferrard2016convolutional} \nocite{kipf2016semi}- \cite{fey2018splinecnn} giving birth to the Graph-based Neural networks. GNNs have shown promising results on representation learning mainly due to the graphs' ability to encode the structure of the data, in contrast to regular grids \cite{shuman2012emerging}, by encapsulating information in the graph's vertices \cite{li2018deeper}. 

Many different graph theoretical domains have made use of the above ideas to this point. Among them, graph classification has attracted a lot of attention lately due to its wide application areas \cite{gomez2017dynamics}, \cite{takerkart2012graph}, \cite{yan2018spatial}. Graph classification is considered as the task to find a mapping $f: G \to T$ that maps each graph in $G$ to a label from a set of target labels $T$. Nonetheless, in graph classification, the continuous-valued vertex attributes, such as the ones that can be found in biological measurements \cite{feragen2013scalable}, are rarely exploited.

In this paper, a flexible multi-view GNN architecture for the task of graph classification is proposed, which propagates both structural and signal information throughout the network. To that end, a novel multi-view graph convolutional layer is presented making use of a hybrid Laplacian, which combines information of the feature space with the structure of the input graph. Subsequently, the use of multiple trainable distance metrics is proposed to cope with training instabilities and overfitting, since it has been proven that single distance metric learning can be prone to overfitting \cite{hoi2010semi}. Finally, by applying a spectral filtering on the graph's signal with different hybrid Laplacians associated to each \enquote{view}, a per \enquote{view} projected signal is calculated. Similar to \cite{su15mvcnn}, a batch normalized max view pooling layer is proposed to aggregate the different per \enquote{view} projected signals. This leads our model to learn from compact generalized representations.

The remainder of this paper is organized as follows: in Section \ref{sec:related}, related work is briefly described. In Section \ref{sec:proposed}, the different components of the proposed methodology are outlined as well as their application to a GNN architecture is detailed. In Section \ref{sec:experimental}, the performed experimental results in several and diverse datasets are reported. Finally, conclusions are drawn in Section \ref{sec:conclusion}.

\section{Related work} \label{sec:related}
Graph convolutional filters can be divided in two main approaches: a) the spatial and b) the spectral ones. Spatial approaches operate directly in the vertices' neighborhoods while spectral approaches operate on the graph's Laplacian eigenspace \cite{zhou2018graph}. 
In PSCN \cite{niepert2016learning}, the authors presented a generalized spatial approach in order to generate local normalized neighborhood regions while in DGCNN \cite{zhang2018end} an end-to-end architecture is proposed that keeps extra vertex information through sorted graph representations from spatial graph convolutions. In order to process graph data with classical CNNs, KGCNN \cite{2017arXiv171010689N} uses a 2-step approach, starting from extracting patches from graphs via common community detection algorithms, embedding them with graph kernels and finally feed them in a classical CNN. Furthermore, another approach proposed in GIN \cite{xu2018powerful} developed an architecture based on Weisfeiler-Lehman test \cite{shervashidze2011weisfeiler}, stating that its discriminating power is equal to the power of the Weisfeiler-Lehman isomorphism test. Finally, in DGK \cite{yanardag2015deep} the authors proposed to leverage the dependency information between sub-structures used in graph kernels by learning their latent representations.

A major breakthrough in graph spectral convolutions was proposed in \cite{defferrard2016convolutional}. Therein, the authors showed the ability of the spectral GNN to extract local features through graph convolutional layers. Moreover, they proposed an efficient pooling strategy on graphs, based on a binary tree structure of the rearranged vertices. Several recently developed graph pooling methods attempt to reduce the number of vertices \cite{ying2018hierarchical}, \cite{defferrard2016convolutional}, \cite{cangea2018towards}, \cite{gao2019graph}, producing coarser and sparser representations of a graph. However, for graph classification, where the number of vertices is relatively small, it is difficult to design such models, while sometimes it could lead to unstable training behaviour \cite{ying2018hierarchical}. Furthermore, the coarsened vertices are not always arranged in a meaningful way. Moreover, graph coarsening processes are usually keeping a fixed number of vertices, which results in training and inference bias \cite{defferrard2016convolutional}. 

There exist various attempts that define pooling operation on graphs. In Graph U-Nets \cite{gao2019graph}, the authors propose a max pooling scheme using a trainable projection vector of the graph signal and then assign the corresponding indices to the adjacency matrix. In DIFFPOOL \cite{ying2018hierarchical}, the authors try to learn the hierarchical structure through a trainable cluster assignment matrix, providing a flexible architecture that can be applied in a plethora of GNNs. In the same direction, CLIQUEPOOL \cite{luzhnica2019clique} also attempts to coarsen the vertices of graph by aggregating maximal cliques. In another recent work called SAGPool \cite{lee2019self}, the authors define an alternative graph pooling method based on self-attention that considers both the vertex features and the graph structure. Finally, in HO-GNN \cite{morris2019weisfeiler}, a GNN architecture is used that takes into account higher-order graph structures at multiple scales.

Some recent approaches have attempted to design architectures that explore different graph structures. In this direction, MGCN \cite{knyazev2018spectral} is developed to capture multi-relational graph relationships through multi-dimensional Chebyshev polynominals. In GCAPS-CNN \cite{verma2018graph}, the authors introduced the notion of capsules in spectral GNNs, in order to encapsulate local higher order statistical information per vertex feature. In a proximal work, CAPS-GNN \cite{xinyi2018capsule} explores additional vertex information by computing hand-crafted statistics. However, both approaches compute extra vertex information explicitly, while they do not involve any trainable components. 

Moreover, one of the first works that introduced single supervised metric learning on graph related tasks is \cite{li2018adaptive}, which is closer to our approach. Nonetheless, this work explores only one graph structure through the graph's Laplacian as opposed to our work where multiple graphs are used. Besides, distance metric learning has also been applied in \cite{ktena2017distance} to resolve the irregular graph matching task, using spectral graph convolutional networks in the field of biomedical applications. Finally, WKPI \cite{zhao2019learning} employ a metric learning scheme, which learns a positive semi-definite weighted-kernel for persistence summaries from labelled data.

In \cite{su15mvcnn} the notion of \enquote{view} was introduced, where each \enquote{view} represents a $2D$ projection of a $3D$ object from a different angle, in order to produce informative representations for the tasks of object recognition and retrieval. In \cite{zhang2018multi}, the authors used the idea of \enquote{views} to construct different brain graphs, where structure was determined by a structural MRI scan and the feature matrices were calculated from multiple tractography algorithms based on diffusion MRI acquisition, to fuse information scattered in different medical imaging modalities for pairwise similarity prediction. However, none of the above view-based approaches introduced a trainable task-driven generation of views, as is the case in our approach.

The majority of graph convolution methods are based on projected graph features into a single graph structure. The motivation behind our approach is that \enquote{views} are artificially created from a common input. These, can extract different learning representations, the same way 3D objects are projected onto 2D images and processed independently. Nevertheless, the fact that the generated \enquote{views} are trainable, as opposed to previous approaches, is a novel and unexplored domain.

\begin{figure}
  \centering
 \includegraphics[width=1\textwidth,height=8cm,keepaspectratio]{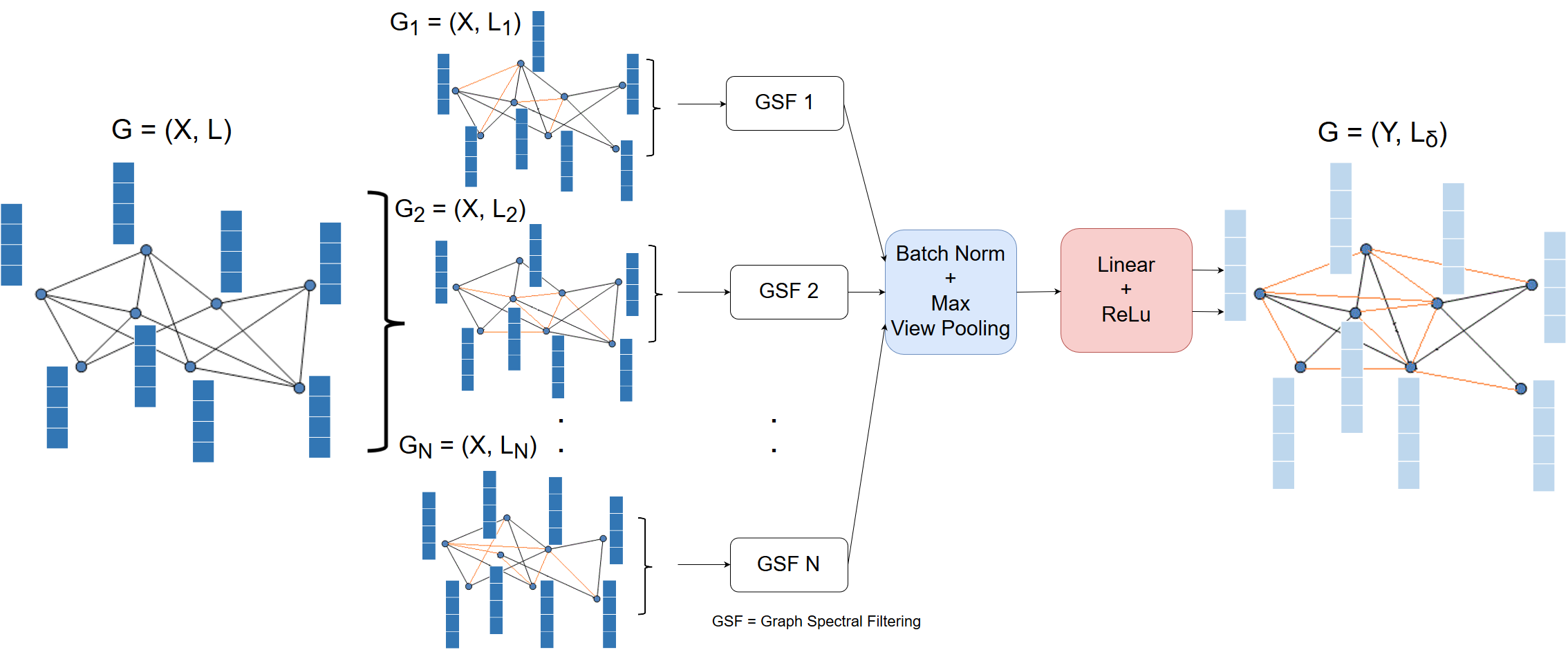}
  \caption{An overview of the proposed layers (multi-view graph convolution and view pooling)}
  \label{fig1}
\end{figure}

\section{Proposed method} \label{sec:proposed}
The proposed work is inspired from \cite{su15mvcnn} and \cite{zhang2018multi} in using the notion of \enquote{views} produced by a shared input (i.e. human vision). The key idea of the proposed multi-view graph convolution (MV-GC) layer is that in each \enquote{view} a different $n$ complete graph ($K_n$) is constructed from the pairwise Mahalanobis distances \cite{de2000mahalanobis} of the vertices in the feature space. Each \enquote{view} is thus encapsulating a different relation between vertices in a distance metric learning context. This information is encoded in the MV-GC layer via the so-called hybrid Laplacian. The hybrid Laplacian is a linear combination of the input graph's Laplacian and a non-Euclidean distance-based Laplacian term, called from now on the \enquote{view} Laplacian, derived from the aforementioned $K_n$ graph. As a result, the spectral graph convolution is approximated by a different hybrid Laplacian eigenspace, producing multiple learnable projections of the same graph signal. Therefore, applying batch normalization as in \cite{ioffe2015batch} and a view-wise max pooling operation, called from now on \enquote{view pooling} layer (VPOOL), an aggregated graph output signal is produced. In contrast to other existing methods \cite{kipf2016semi} that keep the structure of the graph constant throughout the network, in the proposed approach the intrinsic graph's structure is altered during the learning process of the GNN. The whole process can be illustrated in Figure \ref{fig1}.

\subsection{Notations and prerequisites}
To ameliorate the readability of this paper, a summary of the notations and some initial background theory to be used is provided. A graph $G$ is defined as $G = (\mathbf{V}, \mathbf{E})$, where $\mathbf{V}$ is a set of vertices and $\mathbf{E}$ a set of edges, with $e_{ij} \in \mathbf{E}$ defined as $(v_i,v_j)$ and $v_i$, $v_j \in \mathbf{V}$. An alternative representation of a graph of $n$ vertices, taking into account the vertices' feature space can be $G = (\mathbf{A}, \mathbf{X})$, where $\mathbf{A} \in \mathbb{R}^{n \times n}$ is the adjacency matrix and $\mathbf{X}= \left[ \mathbf{x}_1, \mathbf{x}_2, .. , \mathbf{x}_n \right]\in \mathbb{R}^{n \times d} $ is the graph's feature matrix where each row contains the corresponding vertex $d$-dimensional feature vector $\mathbf{x}$. The feature matrix is also called the signal of graph $G$. Consequently, the graph Laplacian is defined as $\mathbf{L} = \mathbf{D} - \mathbf{A}$ where $\mathbf{D} \in R^{n \times n}$ is the diagonal degree matrix with $(\mathbf{D})_{ii} = \sum_{j}(\mathbf{A})_{ij}$. Finally, a commonly used normalized version of the Laplacian operator is defined as follows \cite{chung1997spectral}: 

\begin{equation}
\mathbf{L}_{norm} \triangleq \mathbf{I}_n - \mathbf{D}^{-1/2} \mathbf{A} \mathbf{D}^{-1/2} .
\label{eqn:laplacian-definition}
\end{equation}

$\mathbf{L}_{norm}$ is a real symmetric, positive semi-definite matrix, with a complete set of orthonormal eigenvectors and their associated ordered real non-negative eigenvalues ${\lambda_{i}}$, with $i=[0,n-1]$. In addition, $\mathbf{L}_{norm}$ is utilized in graph spectral convolution, due to the fact that its eigenvalues lie in the range $[0, \lambda_{max} ]$, with $\lambda_{max} \leq 2$ and is widely adopted in most spectral GNN architectures. 

The generalized Mahalanobis distance, with transformation matrix $\mathbf{M}$, for any positive semi-definite matrix and vectors $\mathbf{x}$, $\mathbf{y}$, is defined as: 
\begin{equation}
    d(\mathbf{x},\mathbf{y})  = \sqrt{  \left(\mathbf{x}-\mathbf{y}\right)^{T}\mathbf{M}\left(\mathbf{x}-\mathbf{y}\right)} \quad  \forall  \mathbf{x},\mathbf{y} ,
    \label{eqn:mahalanobis}
\end{equation}
where $^T$ denotes the transpose operation. The above equation can represent a quantitative measure of dissimilarity between vectors $\mathbf{x}$ and $\mathbf{y}$, which are drawn from the same distribution with covariance matrix $\mathbf{C}$ $=\mathbf{M}^{-1}$. For $\mathbf{M} = \mathbf{I}$, eq. \ref{eqn:mahalanobis} reduces to the Euclidean distance. The latter assumes that the variances among different dimensions to be one and covariances to be zero, which is rarely the case in real life applications.

Furthermore, let $diag$ be the operator that maps the main diagonal of an $c \times c$ matrix into a vector $\in \mathbb{R}^c$ and $\mathbf{1}_d$ be a $d$ element vector of ones. A useful equation, derived from the properties of the Hadamard product \cite{horadam2012hadamard}, denoted as $ \odot $, is that for given matrices $\mathbf{A} , \mathbf{B} \in R^{c \times d}$:
\begin{equation}
    diag( \mathbf{A} \mathbf{B}^T) = ( \mathbf{A} \odot \mathbf{B}) \mathbf{1}_d  \in \mathbb{R}^{c} .
    \label{eqn:mahalanobis-property}
\end{equation}


Batch normalization across a set of $N$ representations (or views) of $X_v$, where $ \textbf{X}_v \in R^{n \times d } $, can be defined as an element-wise operation as: 
\begin{equation}
    BN(x_v) = \gamma_{v}(\frac{x_v - \mu(\textbf{X}_v)}{\sigma(\textbf{X}_v)}) + \beta_{v}, 
\end{equation}
where $x_v$ denotes the elements of $\textbf{X}_v$, while $\gamma_v$ and $\beta_v$ are trainable scalars that correspond to a single view. $\mu_v(\textbf{X}_v)$ is the mean value of the elements of matrix $\textbf{X}_v$ and $\sigma$ their standard deviation.

Similarly, given $N$ representations $\textbf{Z} = [ \textbf{X}_1 , \textbf{X}_2, .., \textbf{X}_N ] $ that may correspond to \enquote{views}, max view-wise pooling \cite{su15mvcnn} can be computed by simply taking the maximum across the different views:

\begin{equation}
(\mathbf{Z})_{i,j} = \max( (\mathbf{X}_1)_{i,j} , (\mathbf{X}_2)_{i,j}, .. , (\mathbf{X}_N)_{i,j}) 
\end{equation}

\subsection{Construction of the hybrid Laplacian}
For each graph $G =\left( \mathbf{A}, \mathbf{X} \right)$, $N$ \enquote{views} are initially created. Each \enquote{view} is associated to a feature transformation matrix $\mathbf{M}_{v} \in \mathbb{R}^{d \times d}$ with $v \in [1, .., N]$. Each $\mathbf{M}_{v}$ is a positive semi-definite matrix defined as $\mathbf{M}_{v} = \mathbf{Q}_{v} \mathbf{Q}_v^T \in \mathbb{R}^{d \times d}$, where $\mathbf{Q}_v$ is randomly initialized at the beginning of the learning process. In particular, taking the generalized Mahalanobis distance between feature vectors $\mathbf{x}_i$ and $\mathbf{x}_j$ of the $i_{th}$ and $j_{th}$ vertices respectively, it follows that:
\begin{equation}
    d\left(\mathbf{x}_i,\mathbf{x}_j\right) = \sqrt{  \left(\mathbf{x}_i-\mathbf{x}_j\right)^{T}\mathbf{M}_v\left(\mathbf{x}_i-\mathbf{x}_j\right)} \quad  \forall  \mathbf{x}_i, \mathbf{x}_j .
    \label{eqn:mahalonobis-nodes}
\end{equation}
Moreover, a feature difference matrix $\mathbf{F} \in \mathbb{R}^{c \times d}$, with $c=\frac{1}{2}n(n-1)$, is defined as the matrix of all feature differences between unique pairs of vertices $\left(v_i,v_j\right) \in \mathbf{V}$, without taking into account their connectivity in $\mathbf{E}$. Based on all the above, eq. \ref{eqn:mahalonobis-nodes} can be written, using eq. \ref{eqn:mahalanobis-property} as:

\begin{equation}
\mathbf{d}_{v} = diag(((\mathbf{F} \mathbf{M}_{v} \mathbf{F}^T)\odot \mathbf{I}_c)^{1/2}) = ( \left( \mathbf{F} \mathbf{M}_{v}\odot \mathbf{F} \right) \mathbf{1}_d)^{1/2} \quad \in \mathbb{R}^{c} 
\label{eqn:view-distance}
\end{equation}

The above step is necessary as it significantly reduces the required memory (see Sec. \ref{subsec:complexity-analysis}), and allows the network to be trained in a end-to-end way. During the learning process, it is attempted to learn optimized supervised distance metrics, from the given data, that minimize the GNN's graph classification loss, through backpropagation. The unique distance pairs of $\mathbf{d}_{v}$ are placed back in a \enquote{view} distance matrix $\mathbf{H}_v \in \mathbb{R}^{n \times n}$. The per \enquote{view} similarity matrix, using the Gaussian kernel, is defined as:

\begin{equation}
 \mathbf{S}_v = exp\left(-\mathbf{H}_v/2\sigma^2\right),
\end{equation}

where $\sigma$ is the standard deviation and the hybrid Laplacian, $\mathbf{L}_{h}$ with $h \in [1, .., N]$ , is calculated as:
\begin{equation}
\mathbf{L}_{h} = \mathbf{L}_{in} + \alpha \mathbf{L}_{v},
\label{eqn:hybrid-laplacian}
\end{equation}
with $\mathbf{L}_{v} = \mathbf{I}_{n} - \mathbf{D}^{-1/2} \mathbf{S}_v \mathbf{D}^{-1/2}$ and $\mathbf{L}_{in}$ being the input graph's Laplacian. The contribution of each \enquote{view} is controlled by the scalar value $\alpha$, which is set to 1, to avoid extra hyperpameter tuning. $\mathbf{L}_{v}$ represents a $n$ complete graph ($K_n$) with $c = \frac{1}{2}n(n-1)$ weighted edges based on the feature differences and the trainable distance matrix $\mathbf{M}_v$. Thus, a dense hybrid Laplacian is produced for each \enquote{view}, encoding different relation types between vertices.

For each view v, we compute $L_v$ by taking into account the different trainable $Q_v$, based only on the features of the graph; thus computing different distances. However, since we want to maintain the original graph's structure, we propagate input-graphs’ connectivity by adding $L_{in}$ to $L_v$. Doing so, we also imitate the successful Res-Net architecture that propagates previous layer outputs to subsequent layers (the so-called \enquote{identity connection shortcut}). All the above contribute to faster convergence and encounter for the vanishing gradient problem in the early layers. 

\subsection{Construction of the graph output signal}
As it is described in \cite{defferrard2016convolutional}, convolution of graph signal $\mathbf{X}$ can be defined in the spectral domain and can therefore be approximated by applying a filter $g_{\theta}$ in the eigenvalues of the Laplacian of a graph as:
\begin{equation}
\mathbf{Y} = g_{\theta}({\mathbf{L}})\mathbf{X} = g_{\theta}({\mathbf{U} \mathbf{\Lambda} \mathbf{U}^T}) \mathbf{X} =\mathbf{U} g_{\theta}(\mathbf{\Lambda}) \mathbf{U}^T \mathbf{X} ,
\label{eqn:spectral-decomposition}
\end{equation}
where $\mathbf{U}$ represents the eigenvectors of $\mathbf{L}$ and $\mathbf{\Lambda}$ is a diagonal matrix whose elements are the corresponding eigenvalues. In order to make the spectral filter independent of the graph size and also restrict it in a local graph's region (simulating the CNNs' localization property), the filter is usually modeled as a polynomial function of powers of $\mathbf{\Lambda}$. This expansion can be approximated by using the recurrent Chebyshev expansion to speed up the computations as:
\begin{equation}
g_{\theta}(\mathbf{\Lambda})= \sum_{p=0}^{K-1}{\theta_p \mathbf{\Lambda}^p}= \sum_{p=0}^{K-1}{\theta_p \mathbf{T}_p( \tilde{\mathbf{\Lambda}}) }  
\end{equation}

with $\mathbf{T}_p( \tilde{\mathbf{\Lambda}})= 2 \tilde{\mathbf{\Lambda}} \mathbf{T}_{p-1}( \tilde{\mathbf{\Lambda}}) - \mathbf{T}_{p-2}( \tilde{\mathbf{\Lambda}}) $, $\tilde{\mathbf{\Lambda}}= \frac{2}{\lambda_{max}} \mathbf{\Lambda}-\mathbf{I}$ and $\text{\boldmath$\theta$}\in \mathbb{R}^K $ is the vector of the spectral coefficients. Given a decomposition, $\mathbf{L}^{p}= ( \mathbf{U} \mathbf{\Lambda}  \mathbf{U}^T)^{p} = \mathbf{U} \mathbf{\Lambda}^{p} \mathbf{U}^T$, $g_\theta$ can be parametrized as a polynomial function of $\mathbf{L}$, that can be calculated from a recurrent Chebyshev expansion of order $K$. Using the re-scaled Laplacian $\tilde{\mathbf{L}}_h = \frac{2}{\lambda_{max}} \mathbf{L}_{h} - \mathbf{I}_n $ the computation of decomposition is avoided. Each Laplacian power can be interpreted as expressing the graph constructed from the $p$-hops thus providing the desired localization property. In our approach, such a spectral filter is created for each \enquote{view}, thus, for each hybrid Laplacian. For graph signal $\mathbf{X} \in \mathbb{R}^{ n \times d} $ the projected features $\mathbf{X}_{v} \in \mathbb{R}^{n \times K\cdot d} $, where each $\mathbf{X}_{v}$ represents a different \enquote{view} signal, are calculated as:

\begin{equation}
\mathbf{X}_{v} = g_{\theta} ( \tilde{\mathbf{L}}_{h} )\mathbf{X} =[\tilde{\mathbf{X}}_{0}, \tilde{\mathbf{X}}_{1} , .. ,\tilde{\mathbf{X}}_{K-1}] \mbox{\boldmath$\theta$}_v , 
\label{eqn:signal-projected-reccurent}
\end{equation}

with $\text{\boldmath$\theta$}_{v}= [\theta_0, \theta_1, .. , \theta_{K-1} ]$ are the learnable coefficients shared across vertices in the same \enquote{view} and $\tilde{\mathbf{X}}_{p} = \mathbf{T}_p( \tilde{\mathbf{L}}_{h}) \mathbf{X} = 2\tilde{\mathbf{L}}_{h} \tilde{\mathbf{X}}_{p-1} -\tilde{\mathbf{X}}_{p-2}$. The first two recurrent terms of the polynominal expansion are calculated as: $\tilde{\mathbf{X}}_{0} = \mathbf{X} $ and $\tilde{\mathbf{X}}_{1} = \tilde{\mathbf{L}}_{h} \mathbf{X}$. Thus, the graph signal $\mathbf{X}$ is projected onto the Chebyshev basis $\mathbf{T}_p( \tilde{\mathbf{L}}_{h})$ and concatenated for all orders $p \in [0,K-1]$, similar to \cite{knyazev2018spectral}.

As a result, multiple complex relationships between neighboring vertices from different \enquote{views} are gradually captured. The receptive field is controlled by $K$ and the trainable parameters $\text{\boldmath$\theta$}_{v}$ adjust the contribution of each Chebyshev basis. The majority of spectral convolution methods are based on projected graph features into a single graph structure, including but not limited to Chebyshev approximation. The usage of several trainable hybrid Laplacians, enables spectral filters to capture information in different projected domains. 

For an input graph $G$, $N$ graph signals $\mathbf{Z}_{s} = \left[\mathbf{X}_{1}, \mathbf{X}_{2}, .. , \mathbf{X}_{N} \right]$ are stacked, as well as $N$ hybrid Laplacians $\mathbf{L}_{s} =\left[\mathbf{L}_{1}, \mathbf{L}_{2}, .. , \mathbf{L}_{N} \right]$. Then, batch normalization is applied to $\mathbf{Z}_s$, followed by a view-wise max aggregation step (VPOOL). This way, the VPOOL layer becomes invariant to random initialization of $\mathbf{Q}_v$ thanks to the batch normalization and also fuses information captured from different \enquote{views} due to the aggregation step. The aggregated view-pooled signal has a shared feature matrix, which is then passed to a linear layer followed by a non-linear activation function $\sigma$ and a Dropout layer (DROP) \cite{srivastava2014dropout}. It has to be noted that applying batch normalization to $\mathbf{Z}_s$ the graph signals from the different views have zero-mean and an activation function in the aggregated output can be meaningfully applied. Less literally, all the above procedure can be summarized in the following equation: 

\begin{equation} 
\mathbf{Y} = DROP(\sigma( VPOOL(\mathbf{Z}_s) \mathbf{W} + \mathbf{b} )) ,
\label{eqn:forward-layer}
\end{equation}

where $\mathbf{Y} \in \mathbb{R}^{n \times m}$ is the graph output signal, with $m$ being the number of the output features. The linear layer parameters are $\mathbf{W} \in \mathbb{R}^{K\cdot d \times m}$, $\mathbf{b} \in \mathbb{R}^{m}$. Based on the above, the proposed method can handle a varying number of input vertices, given that the features space is of constant dimension between different graphs (i.e., $d$ is constant between different graphs of the same dataset). In other words, in eq.\ref{eqn:forward-layer} the model parameters are only dependent to the Chebyshev degree $K$ and the feature vector; thus independent of the vertices of the graph.
In addition, the linear layer learns to process aggregated compact signals from multiple views which results in better generalization properties.
Furthermore, based on the indices of VPOOL, the corresponding most frequent hybrid Laplacian is passed to the next layer, which we call \enquote{dominant} hybrid Laplacian, defined as $\mathbf{L}_{\delta}$. Max and mean view pooling have also been tested on $\mathbf{L}_{s}$, giving slightly inferior results.\par
Finally, the overall multi-view GNN architecture consists of three MV-GC layers, each one followed by a VPOOL and a linear layer as depicted in eq. \ref{eqn:forward-layer}. In the first MV-GC layer the input Laplacian is calculated from the adjacency matrix as in eq. \ref{eqn:laplacian-definition}, which we call intrinsic Laplacian. In the rest layers, the \enquote{dominant} Laplacian $\mathbf{L}_{\delta}$ of the previous VPOOL layer is the input Laplacian of the next. The choice of $m$ provides the flexibility to adjust the number of trainable parameters and model complexity. For $m>K\cdot d$ the linear layer behaves as an encoding layer (usually the first layer of our architecture), while for $m<K\cdot d$ acts as a decoding layer. In the last layer, mean and max operators are applied in the graph output signal across the vertex dimension, similar to \cite{cangea2018towards} and \cite{xu2018representation}, which are further concatenated. This step further aggregates the learned features, while concurrently reduces complexity. The concatenated output features are then fed to a 2 fully connected layers. For the prediction, a softmax function is used to produce the final output $\Bar{\mathbf{y}}_{pred}$. The whole network architecture is trained end-to-end. As loss function the cross entropy is used with $q$ classes, given that $\mathbf{y}_T$ is the target one-hot vector, defined as:
\begin{equation}
l_{s} = -\sum_{1}^{q}{ { \mathbf{y}_T}\log( \Bar{\mathbf{y}}_{pred}) . }
\label{eqn:loss-cross-entropy}
\end{equation}

\subsection{Computational complexity analysis} \label{subsec:complexity-analysis}
The proposed method produces dense graph representations. The proposed MV-GC layer requires the calculation of all the differences between vertex features, which is of $O(n^2)$. As shown in eq. \ref{eqn:view-distance}, exploiting distance matrix symmetry, we only used the feature differences between unique pairs of vertices. As a consequence, the time complexity of dense non-square matrix multiplications was reduced from $O(n^2 d^2)$ to $0.5 O(n^2 d^2)$ and Hadamard element-wise multiplication from $O(n^2 d)$ to $0.5O(n^2 d)$ per view. Although time complexity remains quadratic, we were able to scale it down by a factor of $4$. With the current implementation, the time complexity of the MV-GC layer is linearly dependent to the number of views. Nonetheless, the required operations for each view are completely independent and can be implemented in a parallel way. In addition, space complexity required for eq. \ref{eqn:view-distance} is reduced from $O(n^4)$ to $0.5 O(n^2 d)$ per view, using eq. \ref{eqn:mahalanobis-property}. As mentioned in \cite{li2018adaptive}, the required learning space complexity is $O(d^2)$ per view and $O(K\cdot d\cdot m)$ per linear layer, which are independent of the number of vertices. Still, it remains a challenging task to scale our layers for big graph data analysis and further performance optimization is left for future work.

\section{Experimental evaluation} \label{sec:experimental}
\begin{table}
  \caption{Graph classification datasets}
  \label{table:datasets}
  \centering

  \begin{tabular}{l  cccccc}
    \toprule
    Datasets   
   & Graphs   
    &  Classes  
    & \makecell{Mean \\ Number of\\ Vertices}   
    & \makecell{ Mean \\   Number of \\ Edges }   
    & \makecell{ Vertex \\ Labels \\  } 
      & \makecell{ Vertex \\ Attributes \\ Dim}\\
    \midrule

MUTAG  & 188 & 2 & 17.93 & 19.79  & Yes & - \\

PTC-MR   & 344 & 2 & 14.29  & 14.69 & Yes & - \\
\hline
SYNTHIE & 400 & 4  & 95.00 & 172.93 & No & 15 \\

ENZYMES & 600 & 6 & 32.63 & 62.14 & Yes & 18 \\

PROTEINS & 1113 & 2 & 39.06 & 72.82  & Yes & 29 \\

\hline 

IMDB-B  & 1000 & 2  & 19.77 & 96.53  & No & - \\

IMDB-M  & 1500 & 3  & 13.00 & 65.94  & No & - \\

COLLAB  & 5000 & 3 & 74.49 & 2457.78  & No & - \\

\bottomrule
\end{tabular}
  \label{table-dataset-description}
\end{table}

\subsection{Dataset description and modifications}
The proposed model (MV-AGC) is evaluated on four biological datasets and four social network datasets, available at \cite{KKMMN2016}. Each dataset has an arbitrary undirected set of graphs with number of $n$ vertices and binary edges, as well as a categorical graph label $T$. A summary description can be found on Table \ref{table:datasets}. In MUTAG \cite{debnath1991structure}, edges correspond to atom bonds and vertices to atom properties. PTC-MR \cite{toivonen2003statistical} is a dataset of $344$ molecules graphs, where classes indicate carcinogenicity in rats. In ENZYMES \cite{schomburg2004brenda}, each enzyme is a member of one of the Enzyme Commission top level enzyme classes. PROTEINS \cite{borgwardt2005protein} consists of $1113$ graphs modeled with vertices being their secondary structure elements and edges between two vertices exists if they are neighbors in the $3D$ space. COLLAB \cite{yanardag2015structural} is a scientific collaboration dataset with $5K$ graphs of different researchers from three fields of Physics. IMDB-B and IMDB-M \cite{yanardag2015deep} consist of ego-networks of actors that have appeared together in a movie and the goal is to find the movies genre. SYNTHIE \cite{morris2016faster} is a synthetic dataset which consists of $400$ graphs with four classes and each vertex has $15$ continuous-valued attributes. All datasets are publicly available\footnote{https://ls11-www.cs.tu-dortmund.de/staff/morris/graphkerneldatasets}\cite{KKMMN2016}.

\begin{table}
  \caption{Mean accuracies with standard deviations using discrete vertex labels - biological datasets. The reported methods are sorted in chronological order.}
  \label{sample-table1}
  \centering
  \scalebox{0.95}{
  \begin{tabular}{l  cccc}
    \toprule
             & \multicolumn{4}{c}{\textbf{Datasets}}\\
    \cmidrule(r){2-5}
    \textbf{Method} & MUTAG & PTC-MR & ENZYMES & PROTEINS \\
    \midrule
    
WL \cite{shervashidze2011weisfeiler} &  86.0 $\pm$ 1.7  &  61.3 $\pm$ 1.4  & 59.05 $\pm$ 1.05 &  75.6 $\pm$ 0.4  \\

WL-OA \cite{kriege2016valid} & 84.5 $\pm$ 1.7  & 63.6 $\pm$ 1.5 & 59.9 $\pm$ 1.1  &  76.4 $\pm$ 0.4  \\
  \hline

DGK \cite{yanardag2015deep}  &  87.44 $\pm$ 2.72   & 60.08 $\pm$ 2.55   & 53.43 $\pm$ 0.91   &  75.68 $\pm$ 0.54 \\

PSCN \cite{niepert2016learning} &  92.63 $\pm$ 4.21  &  62.29 $\pm$ 5.68 & - & 75.89 $\pm$ 2.76  \\

KGCNN \cite{2017arXiv171010689N} & - & 62.94 $\pm$ 1.69 & 46.35 $\pm$ 0.23 & 75.76 $\pm$ 0.28  \\ 

DGCNN \cite{zhang2018end} & 85.83 $\pm$ 1.66 & 58.59 $\pm$ 2.47 & - & 76.26 $\pm$ 0.24  \\

DIFFPOOL \cite{ying2018hierarchical} & - & - & 62.53 & 76.25  \\
 
GIN \cite{xu2018powerful} & 90.0 $\pm$ 8.8 & 66.6 $\pm$ 6.9 & - & 76.2 $\pm$ 2.6  \\

GCAPS-CNN\cite{verma2018graph} & - & 66.01 $\pm$ 5.91 & 61.83 $\pm$ 5.39 & 76.40 $\pm$ 4.17  \\
 
MGCN \cite{knyazev2018spectral} & 89.1 $\pm$ 1.4  & - & 61.7 $\pm$ 1.3 & 76.5 $\pm$ 0.4  \\ 

Graph U-Nets \cite{gao2019graph}  & - & - & - & 77.68\\

SAGPool \cite{lee2019self} &  - & - & - & 71.86 $\pm$ 0.97 \\

CLIQUEPOOL \cite{luzhnica2019clique} & -  & - & 60.71 & 72.59  \\

 WKPI  \cite{zhao2019learning}   & 88.3 $\pm$ 2.6   & - & - & 78.5 $\pm$ 0.4  \\ 

  \toprule
\textbf{MV-AGC (Ours)}&   \textbf{92.98 $\pm$ 5.12}   &  \textbf{74.45 $\pm$ 3.42} 
 &  \textbf{64.57 $\pm$ 5.27}  & \textbf{78.81 $\pm$ 3.31}   \\ 
\bottomrule
\end{tabular}} 
  \label{table-biological}
\end{table}

Some biological datasets had both discrete vertex labels and continuous-valued vertex attributes. For a fair comparison between other approaches we used the discrete vertex labels in Table \ref{table-biological}. However, classification accuracy can be further improved using the continuous-valued attributes, as shown in Table \ref{table-continuous}. The proposed model is also tested in the aforementioned social network datasets, where vertex information is not provided, using the one hot encoding of the degree of each vertex as feature vector up to a certain degree, in our case from 30 to 50, as in \cite{ying2018hierarchical}. Using the one hot encoding of the discrete vertex labels usually falls into the case of all the vertices in the graph to have the same label, which renders the computation of $\mathbf{L}_{v}$ to be infeasible. We included the mentioned data to train the model apart from the \enquote{view} transformation matrices $\mathbf{M}_v$. An experimental evaluation was conducted, following the conventional approach of 10 fold cross-validation, similar to \cite{knyazev2018spectral}, \cite{yanardag2015deep}, \cite{ying2018hierarchical} and mean classification accuracy across folds with standard deviation are reported in Tables \ref{table-biological}-\ref{table-ablation-study}. The continuous-valued attributes were preprocessed with mean/std normalization before plugged in the network. The datasets are presented in ascending order with respect to their containing number of graphs.

\begin{table}
  \caption{Mean accuracies with standard deviations using vertex degrees - social network datasets}
  \label{sample-table4}
  \centering
 \begin{tabular}{lccc}
     \toprule
             & \multicolumn{3}{c}{\textbf{Datasets}}\\
    \cmidrule(r){2-4}

    \textbf{Method} & IMDB-B & IMDB-M  & COLLAB \\
    \midrule

DGK \cite{yanardag2015deep} &  66.96 $\pm$ 0.56  & 44.55 $\pm$ 0.52   & 73.09 $\pm$ 0.25  \\

KGCNN \cite{2017arXiv171010689N}  &  71.45 $\pm$ 0.15 & 47.46 $\pm$ 0.21  & 74.93 $\pm$ 0.14 \\
 
DGCNN \cite{zhang2018end}   & 70.03 $\pm$ 0.86 & 47.83 $\pm$ 0.85   & 73.76 $\pm$ 0.5 \\
 
PSCN \cite{niepert2016learning}  &  71.00 $\pm$ 2.29 & 45.23 $\pm$ 2.84  & 72.60 $\pm$ 2.15 \\

DIFFPOOL \cite{ying2018hierarchical}   &  - & - &  82.13 \\

GIN \cite{xu2018powerful}   &  75.1 $\pm$ 5.1 & 52.3 $\pm$ 2.8  & 80.6 $\pm$ 1.9 \\

Graph U-Nets \cite{gao2019graph}  & - & - & 77.56 \\

CLIQUEPOOL \cite{luzhnica2019clique}  & - & - & 74.50 \\
 
GCAPS-CNN\cite{verma2018graph}   & 71.69 $\pm$ 3.40 & 48.50 $\pm$ 4.10  & 77.71 $\pm$ 2.51 \\
 
CAPS-GNN \cite{xinyi2018capsule}   & 73.10 $\pm$ 4.83 & 50.27 $\pm$ 2.65  & 79.62 $\pm$ 0.91 \\

HO-GNN \cite{morris2019weisfeiler} & 74.2 & 49.5  & - \\

WKPI \cite{zhao2019learning}   & 75.1 $\pm$ 1.1 & 49.05 $\pm$ 0.4   & - \\
 
 \toprule
\textbf{MV-AGC (Ours)}  &  \textbf{78.20 $\pm$ 3.05} & \textbf{53.47 $\pm$ 3.62}
 &  \textbf{82.41 $\pm$ 1.09} \\

    \bottomrule
  \end{tabular} 
  \label{table-social-network}
\end{table}

\subsection{Implementation details} 
The described multi-view GNN architecture is implemented in the PyTorch \cite{paszke2017automatic} framework. Training is achieved using the stochastic gradient descent optimizer \cite{bottou2010large} with a constant learning rate ranging from $4\cdot e^{-4}$ to $8\cdot e^{-3}$ and single batch size. As non-linear activation function ReLU is used. Dropout was used to avoid overfitting, mostly on the small data sets, similar to \cite{niepert2016learning}. Chebyshev degree $K$ was set to $6$ for all conducted experiments. The hidden fully connected layer has $128$ units. $\mathbf{Q}_v$ is initialized with a random uniform distribution in range $(0,1)$. $\mathbf{M}_v$ is regularized in each iteration divided with it's maximum element, although preserving it's desired properties. The number of views per MV-GC layer is set to $8$ for the first layer and $6$ for the others. The epochs per dataset vary from 30, in COLLAB, to 80, in the ENZYMES dataset. The number of output features $m$ of each linear layer was set to $80$, $128$ and $256$ for the bioinformatics datasets. For the COLLAB dataset, a smaller number of views per MV-GC layer was used, due to high computational demands. For all datasets the experiments were conducted in a NVIDIA GeForce GTX-1080 GPU with $12$GB of memory and $16$ GB of RAM.

An incremental training strategy was adopted to choose the model hyperparameters e.g. number of views. First, we used a single MV-GC layer and we observed that multiple views helped significantly in graph classification. As a sanity check, we inspected the gradients of $Q_v$ that roughly started from a mean value of $10^{-3}$ and reached a mean value of $10^{-8}$ at the end of the training. Then, due to complexity we could not fit more than three layers, which provided the best results. We made hyperparameter tuning and trained the network on two datasets and used the same model and hyperparameters in the other datasets.

\begin{table}
  \caption{Mean accuracies with standard deviations using only the continuous-valued vertex attributes}
  \label{sample-table2}
  \centering
   \begin{tabular}{lccc}
       \toprule
             & \multicolumn{3}{c}{\textbf{Datasets}}\\
    \cmidrule(r){2-4}   
    
    \textbf{Method} & SYNTHIE & ENZYMES & PROTEINS \\
\midrule
HGK \cite{morris2016faster} & 86.27 $\pm$ 0.72  & 66.73 $\pm$ 0.91 & 75.14 $\pm$ 0.47 \\

GraphHopper  \cite{feragen2013scalable} & -  & 69.60 $\pm$ 1.30 & -  \\

SP \cite{borgwardt2005shortest}   & -  & 71.30 $\pm$ 1.30 & 75.50 $\pm$ 0.80  \\

FGW \cite{vayer2018optimal} & - & 71.00 $\pm$ 6.76 & 74.55 $\pm$ 2.74 \\
PROPAK \cite{neumann2016propagation}  & -  & 71.67 $\pm$ 5.63 & 61.34 $\pm$ 4.38  \\

   \toprule
\textbf{MV-AGC (Ours)} & \textbf{90.02 $\pm$ 3.87}  & \textbf{72.62 $\pm$ 2.63} & \textbf{77.66 $\pm$ 2.51}  \\ 
  
    \bottomrule
  \end{tabular} 
  \label{table-continuous}
\end{table}

\subsection{Experimental results}
We evaluate our model compared to other graph deep learning approaches, as well as classical graph kernel methods. As presented in Table \ref{table-biological}, superior results in $4$ biological datasets are reported, using vertex labels as one hot encodings. A significant gain of $7.85$ was reached in PTC-MR dataset, which is justified in the fact that the provided vertex labels of the dataset are of critical importance for classification. In social network datasets, as shown in Table \ref{table-social-network}, using vertices degrees as input features, we report state-of-the-art results in $3$ social network datasets. Our method surpasses all previous approaches in the datasets that contain continuous-valued feature attributes as depicted in Table \ref{table-continuous}. As show in Table \ref{table-ablation-study}, where we vary the number of \enquote{views} per MV-GC layer without any hyperparameter tuning, there is a significant increase in generalization as the number of views per layer increases up to a certain value. Having only one trainable transformation matrix per layer would reduce the model close to \cite{li2018adaptive}, which is usually unstable to train and does not produce generalized representations. There is also a trade-off in the choice of views between training time and accuracy. Nevertheless, Table \ref{table-ablation-study} clearly proves that the idea of multi-view metric learning for graph classification is non-trivial.

\begin{table}
  \caption{Model (MV-AGC) comparison with varying number of views per layer}
  \label{sample-table3}
  \centering
  \begin{tabular}{c   ccc}
       \toprule
             & \multicolumn{3}{c}{\textbf{Datasets}}\\
    \cmidrule(r){2-4}  
    
 \textbf{\# Views } & MUTAG & ENZYMES & PROTEINS \\
\midrule
 2  & 87.71 $\pm$ 6.3  & 56.02 $\pm$ 6.3  & 75.36 $\pm$ 3.9   \\

 3  & 88.88 $\pm$ 5.4  & 57.61 $\pm$ 3.7 & \textbf{78.08 $\pm$ 3.4}  \\

 6  & 90.43 $\pm$ 5.1  & \textbf{62.27 $\pm$ 4.3} & 77.83 $\pm$ 3.7 \\

 9  & \textbf{90.52 $\pm$ 5.4}  & 61.10 $\pm$ 4.6 & 77.81 $\pm$ 3.4  \\

    \bottomrule
  \end{tabular} 
  \label{table-ablation-study}
\end{table}

\subsection{Discussion}
Some limitations that have to be taken account that influence the model's standard deviation and accuracy, based on our experimental study, are the following:
a) the number of samples per dataset, b) the variability in the number of vertices, c) how informative is the graph signal for the prediction, and d) the choice of $\alpha$ in Eq. \ref{eqn:hybrid-laplacian}. It is observed that it is more difficult to estimate optimal global affine transformations as the number of samples increase and as the graphs have a wider range of vertices. On the other hand, in PTC-MR we observed a huge gain, because the graph signal is a significant feature for classification and the vertex variability is low. The choice of $\alpha$ to be equal to $1$ introduces some extra standard deviation, but it still yields better accuracy for the majority of datasets. The reason behind this choice is to not scale down the gradient values on the affine transformation matrices. Moreover, large-scale graphs (with more than 3000 vertices) cannot be process in the referenced hardware. In a future work, we aim to encounter this limitation. 

We claim that MV-AGC best fits continuous-valued datasets, due to the multi-view representation of the feature space. This is demonstrated in Table \ref{table-continuous} by the higher gains in the continuous-valued datasets. As a future work, the convergence of the network with respect to the views will be further investigated.

\section{Conclusions} \label{sec:conclusion}
In the present work, a graph analog of multi-view operations in CNNs was developed. We propose a novel multi-view GNN architecture able to exploit vertices information in an adaptive manner. The proposed MV-GC layer generates \enquote{views} representing multiple graph structures, based on a trainable non-Euclidean distance metric learning process. We explore pairwise feature relationships inside a graph via the newly introduced hybrid Laplacian. Spectral filtering is applied to the input graph signal with a different hybrid Laplacian per \enquote{view}, producing multiple projected signals and, thus, encapsulating different relations between the data. A new view pooling layer is also introduced, able to fuse information from different views. The proposed multi-view graph distance metric learning methodology can also be applied in other graph convolutional schemes, which is out of the scope of this paper and left as a future research direction. Our model (MV-AGC) provides state-of-the-art results in $4$ bioinformatics datasets with discrete vertices labels, as well as in $3$ social network datasets. Finally, the proposed method outperforms previous approaches in all datasets with continuous-valued vertex attributes. 

\section{Acknowledgement}
The work presented in this paper was supported by the European Commission under contract H2020-822601 NADINE.

\bibliographystyle{splncs04}
\bibliography{egbib}
\end{document}